\documentclass[lettersize,journal]{IEEEtran}
\usepackage[caption=false,font=normalsize,labelfont=sf,textfont=sf]{subfig}                             

\usepackage[table]{xcolor} 
\usepackage{amsmath}
\usepackage{amssymb}
\usepackage{mathnotation}
\usepackage{amsthm}
\usepackage{cite}
\usepackage{tabularx}
\usepackage{booktabs}
\usepackage[colorlinks=true, linkcolor=black, citecolor=black, urlcolor=black]{hyperref}
\usepackage{multirow}
\usepackage[ruled,linesnumbered]{algorithm2e}
\usepackage[multiple]{footmisc}
\usepackage{censor}

\theoremstyle{definition}
\newtheorem{definition}{Definition}
\newtheorem{proposition}{Proposition}

\newcommand{\figref}[1]{Fig. \ref{#1}}

\StopCensoring

\title{\LARGE \bf
Stochastic Physics-Informed Neural Networks on Lie Groups for Learning Underwater Vehicle Dynamics}

\author{\censor{Evan F. Palmer}, \censor{Ross L. Hatton}, and \censor{Geoffrey A. Hollinger}
\thanks{*This work was supported in part by \censor{ONR award N00014-23-1-2171}. The work of \censor{Evan Palmer} was supported by \censor{the} \censor{National Defense Science and Engineering Graduate (NDSEG) Fellowship.}}%
\thanks{\censor{Evan F. Palmer}, \censor{Ross L. Hatton}, and \censor{Geoffrey A. Hollinger} are with \censor{the Collaborative Robotics and Intelligent Systems (CoRIS) Institute},
        \censor{Oregon State University, Corvallis OR 97331, USA}
        {\tt\small \censor{\{palmeeva}, \censor{ross.hatton, geoff.hollinger\}@oregonstate.edu}}}%
}

\begin{document}

\maketitle
\thispagestyle{empty}
\pagestyle{empty}

\begin{abstract}

Accurate models of underwater vehicle motion are needed for autonomous execution of marine tasks like infrastructure inspection and scientific sampling. However, such motion is challenging to characterize using traditional physics-based methods. This paper presents a novel data-driven framework for learning stochastic underwater vehicle dynamics. Using Euler-Poincar\'{e} dynamics and the geometry of Lie groups, we develop a stochastic physics-informed neural network architecture that respects the physical and geometric constraints of underwater vehicles. Our approach leverages structure-preserving stochastic integration and builds upon moment matching and finite dimensional matching to ensure geometrically-consistent training. We evaluate our approach in simulation and on an underwater vehicle navigating dock pylons in a harbor environment. The results demonstrate that our method learns accurate and robust dynamics models, enabling safe model-based control in challenging marine environments.


\end{abstract}
\section{Introduction}\label{sec:introduction}

Safely controlling underwater robots requires accurate dynamics models. Dynamics models for underwater systems are traditionally obtained using system identification based on experimental and computational fluid dynamics. However, these methods struggle to fully characterize underwater motion, which hinders model accuracy and expressivity. Data-driven system identification has emerged as a promising alternative to classical system identification. These methods leverage collected observations of a system to learn an approximation of the system's dynamics, reducing manual modeling requirements and improving model accuracy.

\begin{figure}
    \centering
    \includegraphics[width=\linewidth]{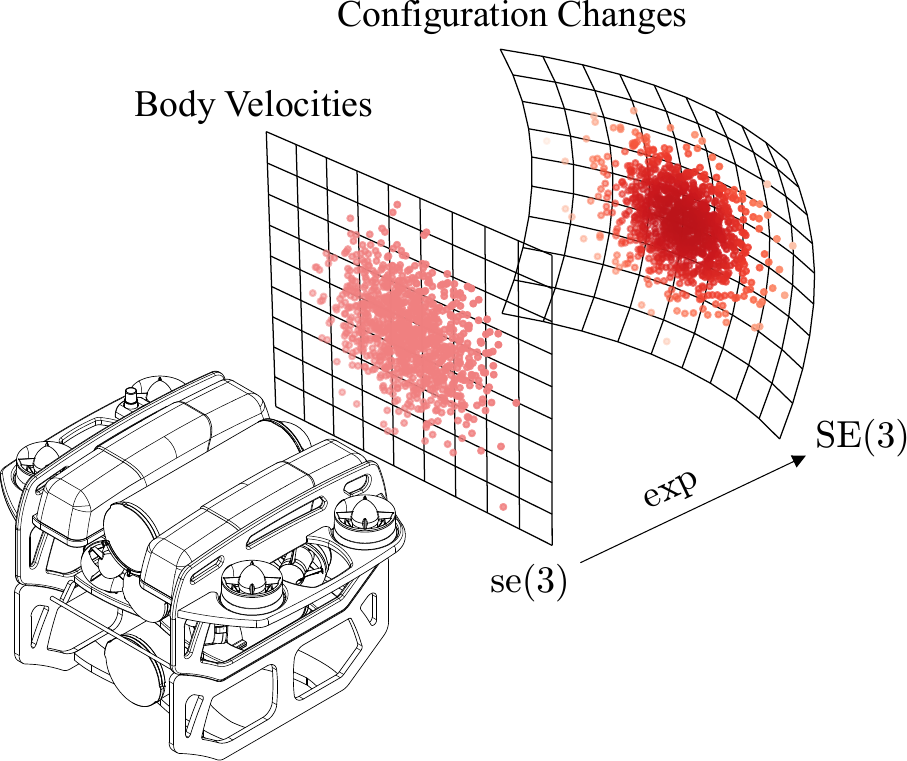}
    \caption{Stochastic physics-informed neural networks on Lie groups learn geometrically-consistent stochastic dynamics by mapping integrated noise defined in the Lie algebra (vector space) to the group (curved space).}
    \label{fig:projection}
\end{figure}

Two main challenges exist in applying data-driven system identification to the underwater domain. First, the complexities involved in deploying underwater robots make it difficult to obtain a sufficient amount of training data for black-box models. Second, black-box models that lack knowledge of the underlying mechanics of a system may be physically inconsistent (e.g., they may not conserve energy in systems that are conservative) \cite{lee2024robot}. Physics-informed neural networks (PINNs) can address these problems by leveraging \textit{a priori} knowledge of a system's dynamics and geometry as inductive bias, improving sampling efficiency and yielding physically-consistent models \cite{djeumou2023autonomous, lutter2023combining, duong2024port, saviolo2022physics}. While PINNs are a promising step forward, these methods are deterministic and do not account for the disturbances that affect underwater vehicles. Feedback control can partially mitigate these effects during execution, but it does not support long-horizon planning or reasoning about a system's motion under uncertainty. Stochastic physics-informed neural networks (SPINNs) can address this gap.

SPINNs extend PINNs by modeling stochastic dynamics using neural networks \cite{kidger2021on}. Prior work has demonstrated that these models are capable of learning the dynamics of robotic systems from limited data and can characterize epistemic and aleatoric uncertainty in the system's motion \cite{djeumou2023how, djeumou2024one, koprulu2025neural}. However, the prior work does not consider the underlying geometry of the system, which can introduce coordinate singularities and degrade trajectory accuracy. This limitation is especially relevant to underwater vehicles because their dynamics evolve on Lie groups.

To address the aforementioned limitations, we propose a method for learning the dynamics of stochastic rigid-body systems whose states evolve on Lie groups. Our approach integrates principles from differential geometry with SPINNs, enabling physically- and geometrically-consistent stochastic dynamics modeling. In summary, our contributions include:
\begin{enumerate}
    \item A novel SPINN architecture for underwater vehicles designed with Euler-Poincar\'{e} dynamics and geometric constraints as inductive biases.
    \item A general method for learning SPINNs on Lie groups using structure-preserving integration and geometrically-consistent training strategies, including moment matching and finite dimensional matching.
    \item Empirical validation of our method in simulation and on an underwater vehicle operating in a harbor environment, showcasing improved predictive performance over baseline methods despite state estimation error and environmental disturbances.
\end{enumerate}
%
Our results demonstrate that respecting the underlying manifold of the system's state space improves SPINN training stability and enables the accurate, uncertainty-aware dynamics models needed for underwater robots to safely operate in oceanic conditions. 
\section{Related Work}
\label{sec:related-work}

The growing complexity of underwater robots and the tasks that they are used for (e.g., underwater docking \cite{vivekanandan2023autonomous} and floating-base manipulation \cite{morgan2022autonomous}) has highlighted the need for accurate models of underwater systems. These models are generally obtained using manual system identification or computational fluid dynamics (CFD) simulations. However, nonlinear hydrodynamic effects are difficult to estimate experimentally and CFD requires simplifying assumptions about the system that hinder accuracy in real-world conditions. 

Data-driven system identification (model learning) addresses these issues by approximating motion models from system data \cite{lee2024robot}, reducing manual modeling effort and errors caused by model mismatch. Examples of data-driven identification in the marine robotics domain include least squares methods \cite{morgan2023probabilistic}, Koopman-based methods \cite{mamakoukas2021derivative, li2022learning, liu2026koopman}, and residual learning using Gaussian processes \cite{amer2025empowering}. More recently, physics-informed neural networks (PINNs), which use \textit{a priori} mechanics knowledge as inductive bias, have been used to improve sample efficiency and consistency with known laws of physics \cite{zhao2024modeling, amer2025modelling, liu2026physics}. However, these works assume that system states evolve in Euclidean space and ignore the underlying symmetries in the dynamics, which preserve relative motion from the perspective of the body frame rather than the world frame. These symmetries dictate that the system state evolves on a Lie group, and disregarding this structure can lead to geometric inconsistencies (e.g., drift from group constraints) and degraded prediction accuracy \cite{duong2024port}. 

Recent work has shown that incorporating the Lie group structure of rigid-body dynamics as an inductive bias in the architecture of a PINN ensures geometrically-consistent predictions and improves predictive performance \cite{wotte2025geometric, duong2024port}. Despite the improvements in geometric PINNs, these models are deterministic and do not capture the underlying stochasticity that can emerge in underwater vehicle dynamics (e.g., tether effects and environmental disturbances). Neglecting these stochastic effects can hinder robustness or prevent successful transfer to field environments where disturbances are abundant \cite{aljalbout2025reality}. Stochastic physics-informed neural networks (SPINNs) could address this problem. 

\begin{figure}
    \centering
    \includegraphics[width=\linewidth]{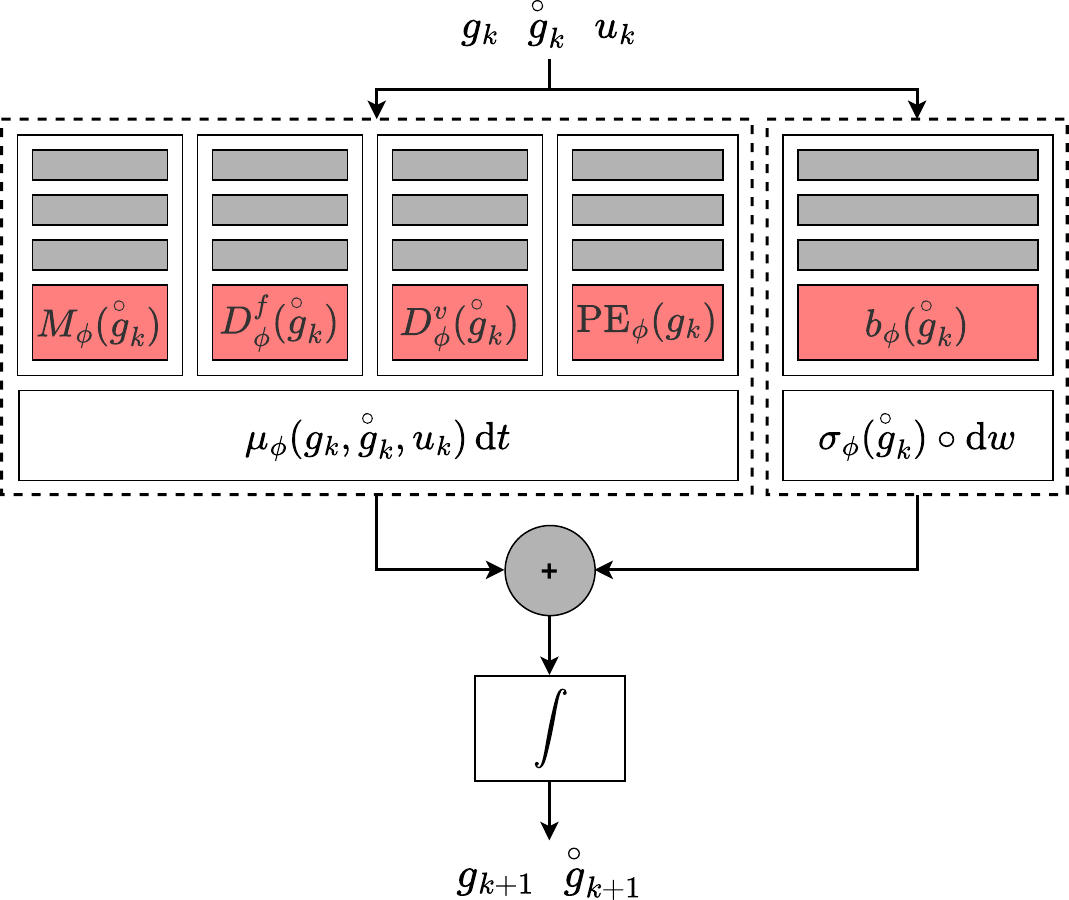}
    \caption{High-level architecture of the geometric stochastic physics-informed neural network. The model approximates rigid-body and fluid-added mass matrix $M$, viscous drag matrix $\viscousdrag$, fluid drag matrix $\fluiddrag$, potential energy $\potentialenergy$, and diffusivity matrix $\covdiffusion$. We parameterize neural networks $M_\neuralparams, \viscousdrag_\neuralparams, \fluiddrag_\neuralparams$, and $\covdiffusion_\neuralparams$ as functions of the body velocity to recognize the system's inherent symmetries. Such symmetry reduction is used to improve data efficiency and is one of the primary motivations for modeling dynamics on Lie groups.}
    \label{fig:architecture}
\end{figure}

SPINNs approximate stochastic differential equations (SDEs) with neural networks, while incorporating prior physics knowledge---such as kinematic constraints \cite{duong2024port, zeng2023latent}, Lagrangian mechanics \cite{lutter2023combining, cranmer2020lagrangian, schultze2026floating}, and Hamiltonian mechanics \cite{greydanus2019hamiltonian, duong2024port}---as inductive biases into the structure of the approximated drift and diffusion. Given a dataset of system state trajectories, the training objective for a SPINN is to optimize the model parameters such that the path space generated by numerically integrating the model matches the observed path space. This objective can be formulated as maximum likelihood estimation \cite{dridi2021learning}, which extends conventional SDE parameter estimation methods \cite{jeisman2006estimation}. If the system dynamics evolve in a latent state space rather than the observation space, the problem can instead be formulated as a variational inference problem \cite{li2020scalable, zeng2023latent}.

There are two limitations to training SPINNs with likelihood-based training strategies. First, the resulting models tend to underestimate the underlying stochasticity in the dynamics \cite{kidger2021on}. Second, standard likelihood-based formulations that assume elliptic diffusion become degenerate when the dynamics are hypoelliptic (i.e., noise enters only a subset of the system state). Djeumou et al. \cite{djeumou2023how} mitigate diffusion collapse and enable epistemic uncertainty estimation by incorporating a distance-aware diffusion loss term into a pseudo-likelihood training objective. This work was extended by \cite{koprulu2025neural}, which adds a second diffusion term to model aleatoric uncertainty. Other work has sidestepped likelihood issues altogether by using distribution-matching techniques, like moment matching \cite{pal2021opening, oleary2022stochastic}, adversarial training \cite{kidger2021gans}, and scoring-based methods \cite{zhang2025efficient, issa2023nonadversarial}. However, these methods represent motion in Euclidean space, which can yield geometrically-inconsistent trajectory distributions for systems whose states evolve on Lie groups. 

Altogether, there is a key gap between stochastic modeling, physical priors, and Lie group structure for learning underwater vehicle dynamics. We address this gap by first integrating Euler-Poincar\'{e} dynamics and Lie group constraints as inductive biases into the architecture of a SPINN. Then, we develop geometrically-consistent training strategies for the SPINN based on moment matching and scoring rules.
\section{Background}


\subsection{Stochastic Physics-Informed Neural Networks}

Define a filtered probability space $\{ \samplespace, \filtration, \filtration_t ,\probmeasure \}$, where $\samplespace$ is the sample space, $\filtration$ is the sigma-algebra, $\filtration_t$ is the filtration, $\probmeasure$ is the probability measure, and we consider the finite time horizon $t \in [0, T]$. Let $\wiener$ be an $m$-dimensional $\filtration_t$-Brownian motion. A (Euclidean) stochastic physics-informed neural network is described by
\begin{equation}
\label{eq:euclidean-spinn}
    \dd \randvar = \neuraldrift(\randvar, \control)\,\dt + \neuraldiffusion(\randvar) \circ \dw,
\end{equation}
where the drift $\neuraldrift : \euclid^n \times \mathcal{U} \rightarrow \euclid^n$ and diffusion $\neuraldiffusion : \euclid^n \rightarrow \euclid^{n \times m}$ are constructed using Lipschitz neural networks parameterized by $\neuralparams$, and $\control$ is a $\mathcal{U}$-valued Markov control strategy \cite{kidger2021on}. Let $\probmeasure_\randvar$ be the probability measure induced by $\randvar$, i.e., its law. Given an observed (data) process $\altrandvar$, the training objective for the SPINN \eqref{eq:euclidean-spinn} is to optimize the model parameters $\neuralparams$ such that $\probmeasure_\randvar$ matches $\probmeasure_\altrandvar$ for some measure of closeness \cite{issa2023nonadversarial}.
\subsection{Scoring Rules}

A well-established class of functionals for measuring closeness between probabilistic forecasts are scoring rules \cite{gneiting2007strictly}. Formally, for probability distributions $\euclidlaw, \alteuclidlaw \in \probmeasure$, a scoring rule $\score{\euclidlaw, \altevent}$ evaluates the quality of the prediction $\euclidlaw$ when the event $\altevent$ is observed. The expected score is its expectation under $\alteuclidlaw$,
\begin{equation}
\label{eq:expected-score}
    \score{\euclidlaw, \alteuclidlaw} = \mathbb{E}_{\altevent \sim \alteuclidlaw}\bigl[\score{\euclidlaw, \altevent} \bigr].
\end{equation}
A scoring rule $\scorerule$ is called proper relative to $\probmeasure$ when no forecast can obtain a score larger than that of the true distribution $\alteuclidlaw$ itself,
\begin{equation}
\label{eq:proper-score}
    \score{\alteuclidlaw, \alteuclidlaw} \geq \score{\euclidlaw, \alteuclidlaw},
\end{equation}
and strictly proper relative to $\probmeasure$ if \eqref{eq:proper-score} holds with equality iff $\euclidlaw = \alteuclidlaw$ \cite{gneiting2007strictly}. A useful example of a proper scoring rule is the kernel score:
\begin{equation}
\label{eq:kernel-score}
    \score{\euclidlaw, \altevent} = \frac{1}{2}\mathbb{E}_{{\event, \event' \sim \euclidlaw}} \bigl[\kappa (\event, \event' ) \bigr] - \mathbb{E}_{\event \sim \euclidlaw} \bigl[ \kappa ( \event, \altevent ) \bigr],
\end{equation}
where $\kappa$ is a continuous negative definite kernel \cite{gneiting2007strictly}. When $\kappa = -\kernel$, with $\kernel$ a positive definite kernel, maximizing the expected score \eqref{eq:expected-score} becomes equivalent to minimizing the squared Maximum Mean Discrepancy (MMD$^2$) between $\euclidlaw$ and $\alteuclidlaw$ \cite{gretton2012kernel}.

\begin{figure}
    \centering
    \includegraphics[width=0.9\linewidth]{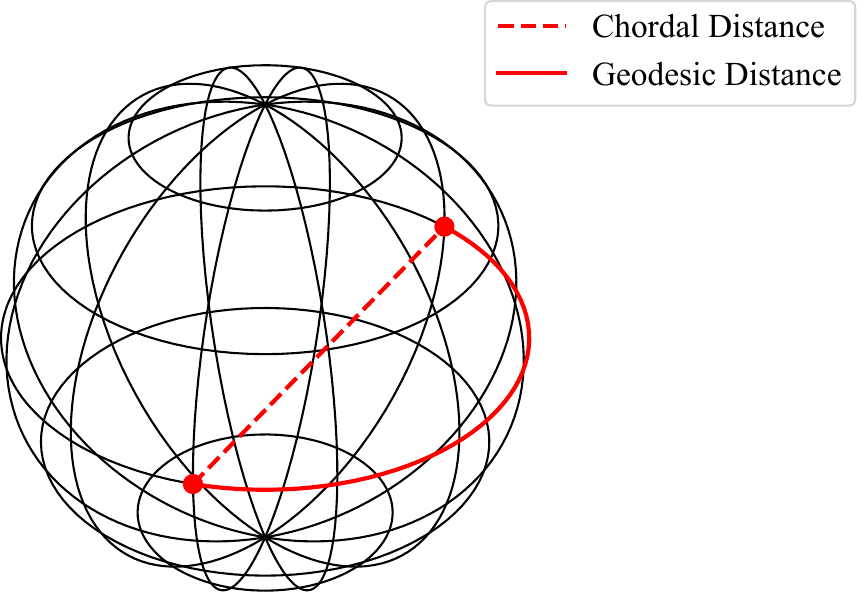}
    \caption{Comparison of the geodesic and chordal distances between two elements on $\SOthree$. The chordal distance measures the straight-line path between two elements in an embedded $\euclid^9$ space, while the geodesic distance measures the shortest arc length along the surface of the manifold. For small angular displacements, the chordal and geodesic distances are approximately proportional to each other. However, because the chordal distance grows more slowly with angular displacement than the geodesic distance (linearly vs. sublinearly), it underestimates true angular displacement. 
    }
    \label{fig:configuration-distances}
\end{figure}
\subsection{Underwater Vehicle Kinematics}

The configuration of an underwater vehicle with respect to a world frame is represented as an element $\fiber$ of the special Euclidean group $\SEthree$,
\begin{equation}
    \SEthree \coloneqq \left\{\begin{bmatrix}R & q \\ 0 & 1\end{bmatrix} \in \euclid^{4 \times 4} \middle| R \in \SOthree, q \in \euclid^3 \right\},
\end{equation}
where $q$ is the position of the system and the orientation $R$ is an element of the special orthogonal group $\SOthree$,
\begin{equation}
    \SOthree \coloneqq \left\{R \in \euclid^{3 \times 3} \middle| \transpose{R}R = \matrixid, \text{det}(R)=1 \right\}.
\end{equation}
$\SEthree$ and $\SOthree$ are examples of Lie groups, i.e., groups in which the underlying set forms a smooth manifold, endowed with a composition operator (which, for these matrix Lie groups, corresponds to matrix multiplication).


The set of velocity vectors that a vehicle can have at a configuration $\fiber$ form a vector space called the tangent space to $\SEthree$ at $\fiber$ denoted $\tans[\fiber]\SEthree$. From the matrix perspective, this vector space contains all of the infinitesimal matrices that can be added to an $\SEthree$ matrix and produce another $\SEthree$ matrix. The tangent space at the group identity $\groupid = \matrixid$ is called the Lie algebra $\sethree$, in which we express the body velocity of the system $\fibercirc$ describing its motion in a body-aligned frame.\footnote{We adopt the \textit{open-circle} notation used in \cite{yang2025geometric}, in which the open circle indicates derivatives taken with respect to the right group-action. An open circle in the overset position $\groupderiv{x}$ denotes tangent vectors (e.g., body velocity $\fibercirc$), whereas the open circle in the underset position $\groupcov{x}$ denotes cotangent vectors (e.g., body momentum $\genmomcirc$ and body force $\lieforce$).}

Velocities in $\tans[\fiber]\SEthree$ are mapped to the Lie algebra by the left trivialization
\begin{align}
\label{eq:body-velocity-increment}
    \inv{\fiber}\dd\fiber = \fibercirc\,\dt \in \sethree,
\end{align}
which describes infinitesimal changes in configuration in the body frame. Similarly, left-translating the velocity increment maps body-frame motion into a configuration increment
\begin{equation}
\label{eq:kinematics}
    \dd\fiber = \fiber\fibercirc\,\dt.
\end{equation}

Points in $\SEthree$ and $\sethree$ are related through the exponential and logarithm maps. Specifically, the exponential map $\exp: \sethree \mapsto \SEthree$ maps a velocity in the Lie algebra to the configuration that would be reached by flowing with that velocity for unit time, while the logarithm map $\log: \SEthree \mapsto \sethree$ performs the inverse operation. These maps are bijective in a neighborhood of the group identity $\groupid$: for rotations smaller than a half-circle rotation, $\log(\exp(\fibercirc)) = \fibercirc$.
\subsection{Underwater Vehicle Dynamics}
\label{subsec:underwater-dynamics}

The Lagrangian $\lagrangian$ of a holonomic underwater vehicle is defined as
\begin{subequations}
\begin{align}
  \lagrangian(\fiber, \fibercirc) &= \kineticenergy(\fiber, \fibercirc) - \potentialenergy(\fiber) \\
  \kineticenergy(\fiber, \fibercirc) &= \frac{1}{2}\transpose{\fibercirc}\metric\fibercirc
\end{align}    
\end{subequations}
with kinetic energy $\kineticenergy$; potential energy $\potentialenergy$; and the symmetric, positive definite mass matrix $M$ (including rigid-body and fluid-added mass effects). Let $\{ \samplespace, \filtration, \filtration_t, \probmeasure \}$ be a filtered probability space on which is defined a $\filtration_t$-Brownian motion $\covwiener$ expressed in the body frame. Non-conservative forces and moments acting on the vehicle can be encoded by applying the Lagrange-d'Alembert principle:
\begin{equation}
\label{eq:lda}
    \delta \int_0^T\!\lagrangian(\fiber, \fibercirc)\;\dt + \int_0^T\!\covprod{\littlelieforce}{\variation}\;\dt + \int_0^T\!\covprod{\covdiffusion(\fibercirc) \circ \covdw}{\variation} = 0,
\end{equation}
where the non-conservative, deterministic forces and moments
\begin{equation}
    \littlelieforce = \control - \viscousdrag \fibercirc - \fluiddrag \bigl( | \fibercirc | \odot \fibercirc \bigr),
\end{equation}
comprise the system input $\control$, viscous drag matrix $\viscousdrag$, and fluid drag matrix $\fluiddrag$ (with $\odot$ the Hadamard product). The diffusion $\covdiffusion(\fibercirc)$ scales the magnitude and direction of non-conservative stochastic effects. Altogether, \eqref{eq:lda} consists of two Lebesgue integrals with respect to the Lebesgue measure $\dt$ and one Stratonovich stochastic integral with respect to $\covdw$.\footnote{We choose the Stratonovich interpretation of the stochastic integral to preserve the standard chain rule of calculus.}

The continuous infinitesimal variations ($\delta \fiber$ and $\delta \fibercirc$) satisfy
\begin{equation}
\label{eq:variations}
    \delta \fiber = \fiber \variation, \qquad \delta \fibercirc = \sddt \variation + \adj_{\fibercirc}\variation,
\end{equation}
where $\variation$ vanishes at the endpoints but is otherwise arbitrary. Substituting \eqref{eq:variations} into \eqref{eq:lda} yields the stochastic Euler-Poincar\'{e} equation
\begin{subequations}
\label{eq:continuous-dynamics}
\begin{align}
    \dd \genmomcirc &= \Bigl(\dadj_{\fibercirc} \genmomcirc + \lieforce \Bigr) \dt + \covdiffusion(\fibercirc) \circ \covdw, \\[3pt]
    \lieforce &= \fiberstar \frac{\dd \lagrangian(\fiber, \fibercirc)}{\dd \fiber} + \littlelieforce,
\end{align}
\end{subequations}
where $\genmomcirc = \metric \fibercirc$ corresponds to the body momentum, $\dadj_{\fibercirc} \genmomcirc$ denotes the rate at which a body moving with constant momentum relative to a fixed spatial observer sees this momentum as changing when observed within its body frame, and $\fiberstar$ pulls spatial forces and torques (e.g., gravitational and buoyant forces) back into the body frame. Expressing \eqref{eq:continuous-dynamics} in terms of the body velocity evolution yields,
\begin{equation}
\label{eq:velocity-dynamics}
    \dd \fibercirc = \inv{M} \Bigl(\dadj_{\fibercirc} \genmomcirc + \lieforce \Bigr) \dt + \metricdiffusion(\fibercirc) \circ \covdw,
\end{equation}
where $\metricdiffusion(\fibercirc) = \inv{M}b(\fibercirc)$ is the cometric-scaled diffusion.

The resulting SDE satisfies the Markov property and defines a hypoelliptic diffusion process, i.e., the noise covariance is rank-deficient in the full system state $(\fiber, \fibercirc)$. This modeling choice reflects the physical reality that stochastic disturbances enter the system as forces and moments rather than direct perturbations on the vehicle's configuration. 

\begin{figure}
    \centering
    \includegraphics[width=\linewidth]{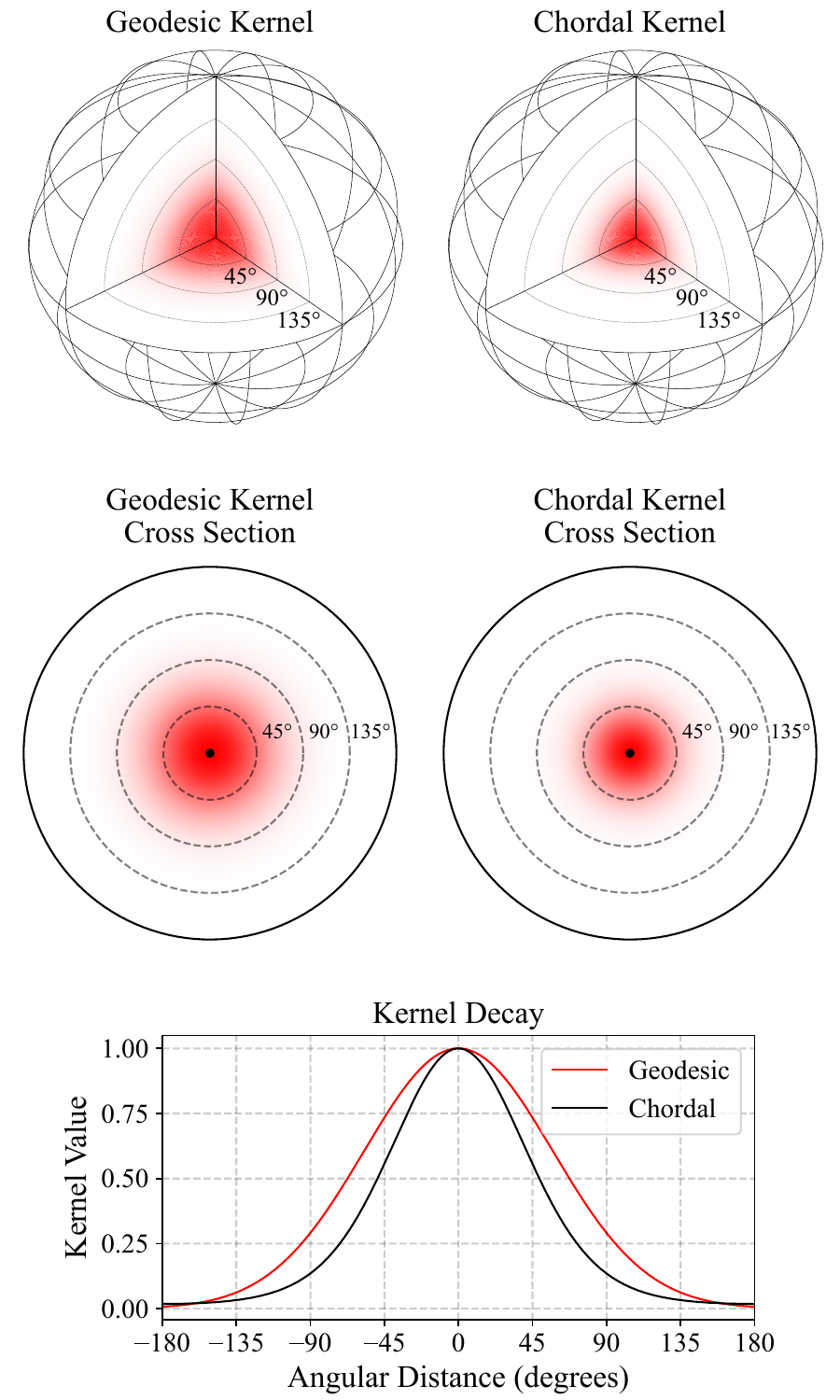}
    \caption{The squared exponential kernel with bandwidth $\gamma = 1$ on $\SOthree$ constructed using the geodesic and chordal distances. Radial distance from the center indicates angular separation from the reference rotation. The faster decay of the chordal kernel reflects the metric stretching of angular distances induced by the Frobenius norm in the embedding space $\euclid^9$.}
    \label{fig:kernels}
\end{figure}

The reader is referred to the following textbooks \cite{marsden1994introduction, chirikjian2012stochastic} for a more detailed exposition of geometric modeling and to \cite{handbook2011fossen} for further reading on marine craft hydrodynamics.

\section{Method}

Define a control scenario as
\begin{equation}
    s = \bigl( \altfiber_0, \altfibercirc_0, u_{1:N}, t_{1:N} \bigr)
\end{equation}
where $\bigl(\altfiber_0, \altfibercirc_0\bigr) \in \SEthree \times \sethree$ is the $\filtration_0$-measurable initial state, $u_{1:N}$ is a sequence of $N$ discrete control inputs, and $t_{1:N}$ are the corresponding timestamps. Given a dataset
\begin{equation}
    \mathcal{D} = \left\{\left(s^j,  \left\{ \altfiber_{1:N}^{j,\,k}, \altfibercirc_{1:N}^{j,\,k} \right\}_{k=1}^{K} \right) \right\}_{j=1}^{J},
\end{equation}
comprising $J$ control scenarios and $K$ state trajectories per scenario, we seek to learn a SPINN whose law $\jointlaw$ captures intrinsic stochasticity in the system dynamics while satisfying the geometric and dynamics constraints of an underwater vehicle. To achieve this, we first propose a novel SPINN architecture for underwater vehicle dynamics. Then, we design two training strategies that extend moment matching and score-based training strategies, respectively.
\subsection{Stochastic Physics-Informed Neural Networks on Lie Groups}
\label{subsec:architecture}

We integrate the kinematic \eqref{eq:kinematics} and Euler-Poincar\'{e} \eqref{eq:velocity-dynamics} equations as inductive biases into the structure of a SPINN to enforce consistency with the dynamics of an underwater vehicle. Let
\begin{subequations}
\begin{align}
\label{eq:hydrodynamic-networks}
    M_\neuralparams &: \sethree \rightarrow \Spdsix, \\
    \viscousdrag_\neuralparams &: \sethree \rightarrow \Spdsix \\
    \fluiddrag_\neuralparams &: \sethree \rightarrow \Spdsix, \\
    \potentialenergy_\neuralparams &: \SEthree \rightarrow \euclid, \\
    \covdiffusion_\neuralparams &: \sethree \rightarrow \text{Diag}_+^6,
\end{align}
\end{subequations}
where $M_\neuralparams$, $\viscousdrag_\neuralparams$, $\fluiddrag_\neuralparams$, $\potentialenergy_\neuralparams$, and $\covdiffusion_\neuralparams$ are Lipschitz neural networks parameterized collectively by $\neuralparams$, approximating the rigid-body and fluid-added mass matrix, viscous drag matrix, fluid drag matrix, potential energy, and diffusivity matrix, respectively.\footnote{Although $M_\neuralparams$ is written as a function of the body velocity $\fibercirc$, this reflects the property that the network is trained using a time sequence of observed body velocities, rather than a model structure that anticipates that the mass will be velocity dependent. In practice, training recovers a constant mass matrix, so its prediction is independent of $\fibercirc$ at evaluation. Constancy can be enforced explicitly by adding a regularization term that penalizes the velocity dependence of $M_\neuralparams$, ensuring that the Euler-Poincar\'{e} constraints remain valid.}$^,$\footnote{The symmetric, positive definite matrix structure is achieved by applying a positive-valued nonlinear function (e.g., softplus, exp, etc.) to the eigenvalues of the $\euclid^{6 \times 6}$ matrices predicted by an MLP.}$^,$\footnotemark\ The resulting learned drift is described by
\begin{equation}
    \neuraldrift(\fiber, \fibercirc, \control) = \begin{bmatrix}
        \fiber \fibercirc \\[3pt]
        \inv{M}_\neuralparams(\fibercirc) \left( \dadj_{\fibercirc} \genmomcirc + \lieforce_\neuralparams(\fiber, \fibercirc, \control) \right)
    \end{bmatrix}
\end{equation}
and the learned diffusion is
\begin{equation}
    \neuraldiffusion(\fibercirc) = \begin{bmatrix}
        0 \\[5pt]
        \neuralmetricdiffusion(\fibercirc)
    \end{bmatrix},
\end{equation}
where we have used Leibniz's rule to write the approximated dynamics in terms of the body velocity evolution. A summary of the model architecture is depicted in \figref{fig:architecture}. The geometric SPINN can be efficiently sampled using a partitioned Lie group integrator, where we use a Lie-Euler step \cite{iserles2000lie} for the configuration evolution and an Euler-Heun step \cite{roberts2012modify} for the velocity evolution. See Appendix~\ref{app:integrator} for additional details.

\begin{figure}
    \centering
    \includegraphics[width=\linewidth]{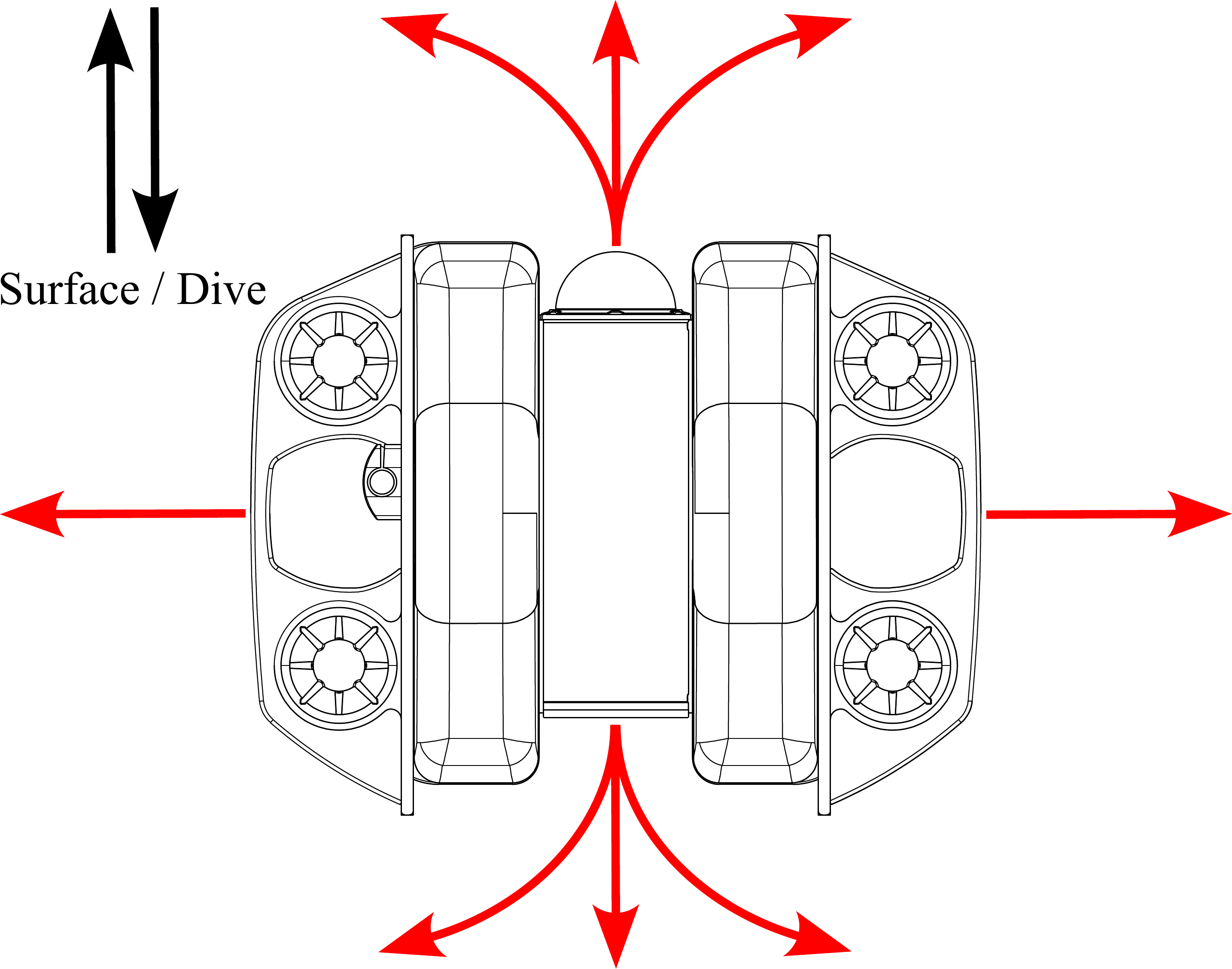}
    \caption{We construct our simulated and ocean training datasets by applying ten feedforward control maneuvers representing typical control actions performed by an underwater vehicle.}
    \label{fig:maneuvers}
\end{figure}
\subsection{Training Geometric SPINNs with Moment Matching}

We establish a training baseline for the geometric SPINN via a naive moment matching strategy.
Given a control scenario $s \in \mathcal{D}$, moment matching minimizes the discrepancy between the mean and covariance (i.e., the first two empirical moments) of the observed sample trajectories $\{\altfiber_{1:N}^k, \altfibercirc_{1:N}^k \}_{k=1}^K$ and the predicted trajectories
\footnotetext{The potential energy is modeled as $\potentialenergy_\neuralparams : \fiber \mapsto \lambda~\text{tanh}\left(\text{MLP}_{\neuralparams_\potentialenergy}(\fiber)\right)$, where the learned scalar $\lambda \in \euclid$ (part of $\neuralparams$) and final tanh nonlinearity are applied to the MLP output to constrain the rate of change in the hidden state \cite{kidger2021on}. We found that without these constraints, the optimizer tends to over-rely on the potential energy term to minimize the loss and implicitly regularizes the hydrodynamic coefficient networks.}
\begin{equation}
    \left\{ \fiber_{1:N}^p, \fibercirc_{1:N}^p \right\}_{p=1}^{P} = \texttt{solve\_sde}(s, \neuraldrift, \neuraldiffusion),
\end{equation}
where $\texttt{solve\_sde}$ represents the Lie group integrator described in Section \ref{subsec:architecture}. We compute the configuration moments at each time step using an iterative procedure that estimates the group-theoretic mean and covariance under the left-invariant distance function for $\SEthree$ \cite{ackerman2013probabilistic}
\begin{subequations}
\begin{align}
    \meanfiber_n &= \arg\min_{c}\hspace{2mm}\sum_{p=1}^{P}\| \log(\inv{c}_n \fiber_n^p) \|^2, \\
    \covariance^{\fiber}_n &= \frac{1}{P}\sum_{p=1}^P \log(\inv{\meanfiber}_n \fiber_n^p)\transpose{\log(\inv{\meanfiber}_n \fiber_n^p)},
\end{align}
\end{subequations}
The velocity statistics are calculated using standard vector methods, which implicitly assume a Euclidean metric on the velocity space:
\begin{subequations}
\begin{align}
    \meanfibercirc_n &= \frac{1}{P} \sum_{p=1}^{P} \fibercirc_n^p, \\
    \covariance^{\fibercirc}_n &= \frac{1}{P} \sum_{p=1}^{P} (\fibercirc_n^p - \meanfibercirc_n)\transpose{(\fibercirc_n^p - \meanfibercirc_n)}.
\end{align}
\end{subequations}
\begin{figure*}
    \centering
    \includegraphics[width=\linewidth]{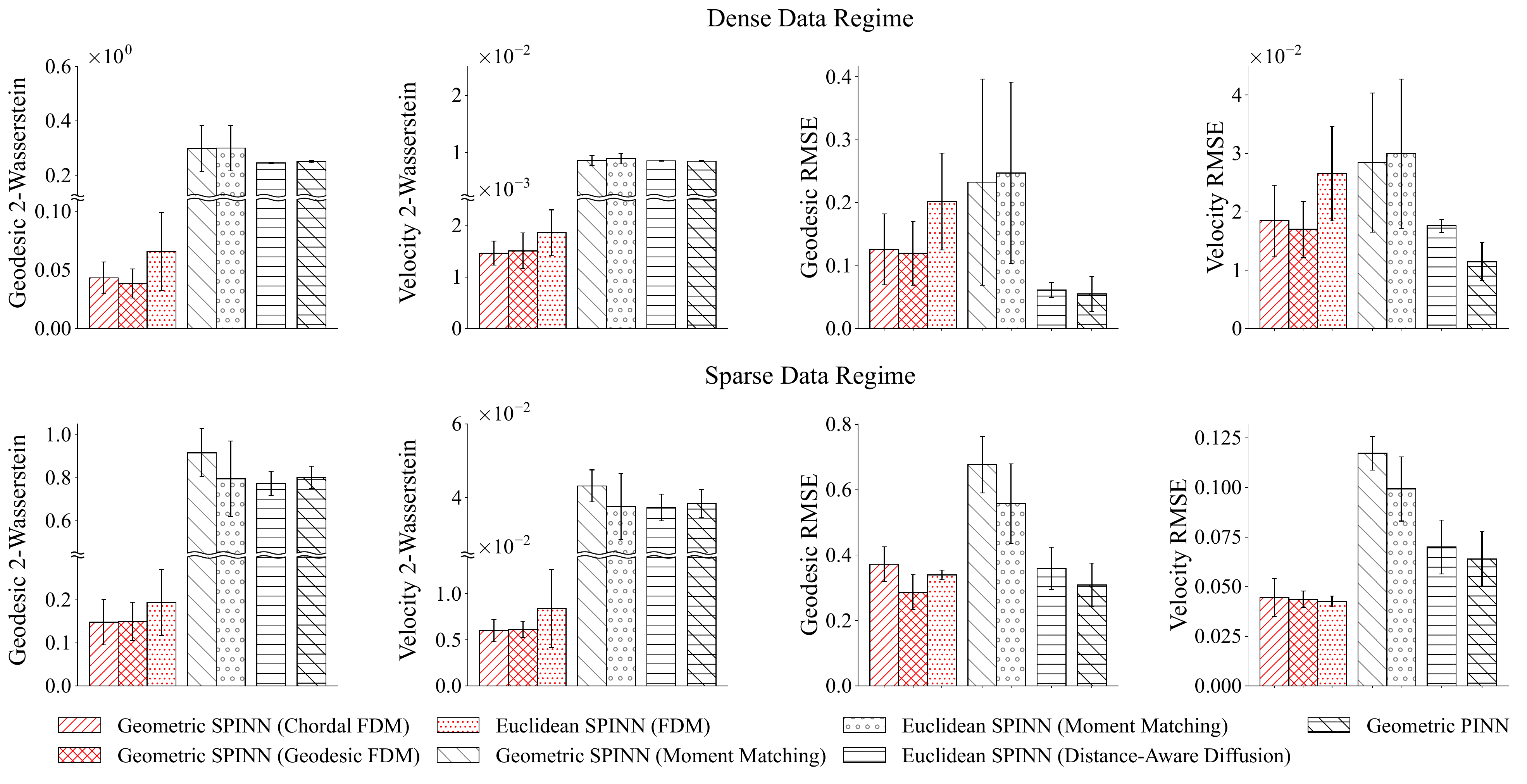}
    \caption{Simulation performance comparison evaluated over 350 samples from each control scenario under the true and learned dynamics. The proposed geometric SPINNs trained via finite dimensional matching (FDM) outperform all baselines in distribution matching while delivering competitive or superior RMSE. Although the Euclidean SPINN trained under FDM achieves competitive distribution matching, it suffers from training instability in the sparse data regime due to the discontinuous state space, converging successfully on only three of five seeds. Conversely, naive training strategies like moment matching fail to capture aleatoric uncertainty and collapse to a near-deterministic solution, regardless of geometric architectural constraints.}
    \label{fig:simulation-results}
\end{figure*}
This approach is repeated for the data moments.\footnote{Uncertainty propagation methods could be used to calculate the moments instead of the Monte Carlo approach used in this work \cite{oleary2022stochastic, barfoot2024state}. We use the Monte Carlo approach because of its functional simplicity.} We define the first-moment loss as
\begin{equation}
    \loss_{\text{mean}}(\neuralparams) = \sum_{n=1}^N \bigl\| \log(\inv{\meanfiber}_n \altmeanfiber_n) \bigr\|^2 + \bigl\| \altmeanfibercirc_n - \meanfibercirc_n \bigr\|^2,
\end{equation}
and the second-moment loss as
\begin{equation}
    \loss_{\text{cov}}(\neuralparams) = \sum_{n=1}^N \bigl\| \covariance^\altfiber_n - \covariance^\fiber_n \bigr\|_F^2 + \bigl\| \covariance^{\altfibercirc}_n - \covariance^{\fibercirc}_n \bigr\|_F^2.
\end{equation}
The resulting moment matching training objective is described by
\begin{equation}
\label{eq:moment-matching}
    \loss_{\text{mm}}(\neuralparams) = \mathbb{E}_{s \sim \mathcal{D}} \left[ \loss_{\text{mean}}(\neuralparams) + \loss_{\text{cov}}(\neuralparams) \right].
\end{equation}
\subsection{Training Geometric SPINNs with Finite Dimensional Matching}

While moment matching offers a straightforward implementation strategy for initial validation, its efficacy is limited by the truncation of higher-order moments, leaving the trajectory distribution underdetermined. We address this limitation with finite dimensional matching (FDM) using scoring rules, which captures the full system dynamics by matching the distribution of state transitions between pairs of time points \cite{zhang2025efficient}. To extend this framework to systems with states evolving on Lie groups, we first define the general matching objective on an arbitrary state space, then instantiate it by constructing kernel scoring rules for Lie group states.

To formalize the general objective, define a filtered probability space $\{\samplespace, \filtration, \filtration_t, \probmeasure \}$. Let $\randvar$ and $\altrandvar$ be predicted and observed $\filtration_t$-adapted Markov processes on $[0, T]$, respectively, taking values in the Polish space $\polish$, endowed with its Borel sigma-algebra $\borel(\polish)$. Given two times $t_1, t_2$ sampled from the uniform distribution $U([0, T]^2)$, we define the joint marginal distribution $\euclidjointlaw$ as the probability measure on the product space $\polish \times \polish$ induced by $\jointvar \coloneqq (\randvar_{t_1}, \randvar_{t_2})$. The product space is equipped with the product sigma-algebra $\borel(\polish) \times \borel(\polish)$. 

Let $\scorerule$ be a kernel scoring rule \cite{gneiting2007strictly} defined on $\polish \times \polish$
\begin{equation}
\label{eq:kernel-score}
    \score{\euclidjointlaw, \altjointvar} = -\frac{1}{2}\mathbb{E}_{\jointvar, \jointvar'}\left[ \kernel(\jointvar, \jointvar') \right] + \mathbb{E}_{\jointvar}\left[ \kernel(\jointvar, \altjointvar) \right]
\end{equation}
where $\jointvar, \jointvar' \sim \euclidjointlaw$, $\altjointvar \coloneqq (\altrandvar_{t_1}, \altrandvar_{t_2})$, and $\altjointvar \sim \alteuclidjointlaw$. If the kernel $\kernel$ is a characteristic kernel \cite{sriperumbudur2010hilbert}, then \eqref{eq:kernel-score} is strictly proper. A scoring rule for $\polish$-valued Markov processes is defined as the expectation of $\scorerule$ under $U([0, T]^2)$
\begin{equation}
    \twoscore{\euclidlaw, \altrandvar} = \mathbb{E}_{t_1, t_2} \bigl[ \score{\euclidjointlaw, \altjointvar} \bigr].
\end{equation}
The expectation of $\twoscorerule$ under $\alteuclidlaw$ is
\begin{equation}
\label{eq:two-time-expectation}
    \twoscore{\euclidlaw, \alteuclidlaw} = \mathbb{E}_{\altevent \sim \alteuclidlaw} \left[ \twoscore{\euclidlaw, \altevent} \right].
\end{equation}
Zhang et al. \cite{zhang2025efficient} showed that if $\scorerule$ is strictly proper, then so is $\twoscorerule$. Therefore, maximizing $\twoscore{\euclidlaw, \alteuclidlaw}$ ensures $\euclidlaw = \alteuclidlaw$, allowing an unbiased estimator of the expected score to serve as a training objective for SPINNs \cite{zhang2025efficient}. Intuitively, this corresponds to minimizing the $\text{MMD}^2$ on the transitions between time points, which captures the dynamics of the system. We now define characteristic kernels for the velocity and configuration joint marginals.

\begin{figure*}
    \centering
    \includegraphics[width=\linewidth]{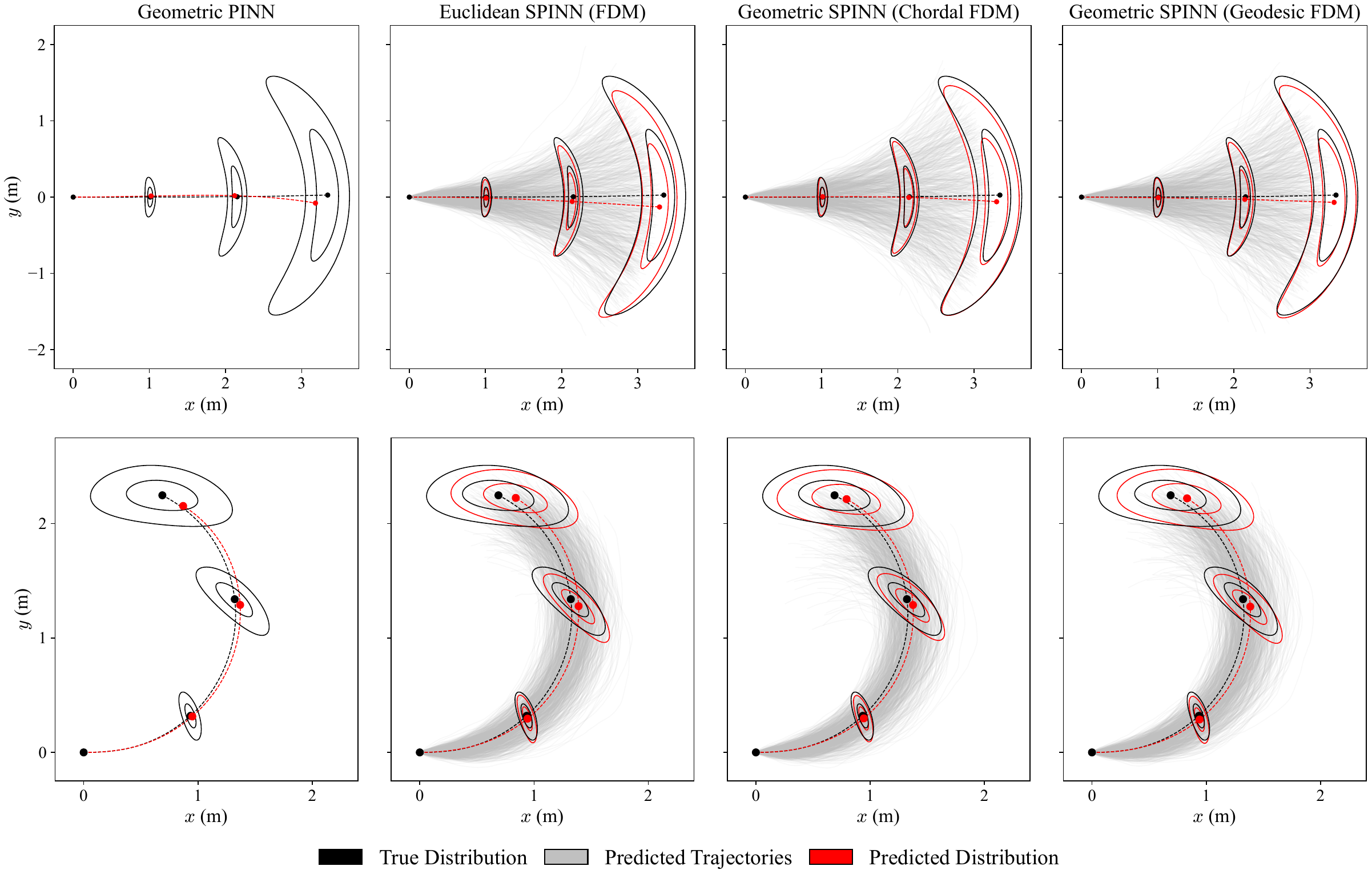}
    \caption{Predicted and true trajectory distributions estimated using 750 samples in the sparse data regime.}
    \label{fig:simulation-distributions}
\end{figure*}

\subsubsection{Velocity Kernel}

The velocity dynamics evolve within a vector space, so the joint marginal distribution $\jointfibercirclaw$ induced by $\jointfibercirc \coloneqq (\fibercirc_{t_1}, \fibercirc_{t_2})$ is supported on the linear product domain $\sethree \times \sethree$. We define the velocity kernel as the product of two automatic relevance determination (ARD) kernels,
\begin{equation}
\label{eq:velocity-kernel}
    \velkernel \Bigl( \jointfibercirc, \altjointfibercirc \Bigr) = \prod_{k \in \{1, 2\}} \exp\left( -\frac{1}{2}\left\| \altfibercirc_{t_k} - \fibercirc_{t_k} \right\|_{\inv{\Gamma}_k}^2\right),
\end{equation}
where $\Gamma_k = \text{diag}(\gamma_k)$ is a diagonal matrix of positive bandwidths $\gamma_k \in \euclid_+^6$, acting as a Riemannian metric on the space. Because $\velkernel$ is equivalent to the product of standard Gaussian radial basis function (RBF) kernels over a Euclidean domain, it is characteristic \cite{sriperumbudur2010hilbert} and the resulting kernel score is thus strictly proper.

\subsubsection{Configuration Kernel}

Choosing a configuration kernel is more nuanced than in the velocity case. Because $\SEthree$ is a non-compact, non-abelian Lie group, recent work on stationary Mat\'{e}rn kernels for Lie groups \cite{azangulov2024stationary1, azangulov2024stationary2} does not apply. To the best of our knowledge, developing a globally-characteristic and computationally-tractable kernel for the Special Euclidean group is an open challenge. We address this problem via two alternatives: (1) altering our notion of distance by utilizing an extrinsic embedding and (2) identifying conditions under which a configuration kernel behaves locally as characteristic.

Let $\jointfiberlaw$ be the joint marginal induced by $\jointfiber \coloneqq (\fiber_{t_1}, \fiber_{t_2})$. We propose two configuration kernels: $\chordalkernel$ and $\geodesickernel$ (see \figref{fig:kernels}). The first kernel $\chordalkernel$ constructs an exponential kernel from the chordal distance,
\begin{equation}
    \chordalkernel \left( \jointfiber, \altjointfiber \right) = \prod_{k \in \{1, 2\}} \exp\left( -\frac{1}{2}\left\| \altfiber_{t_k} - \fiber_{t_k} \right\|_{\inv{\Gamma}_k}^2\right),
\end{equation}
with $\gamma_k \in \euclid_+^{12}$. By embedding $\SEthree$ elements into a flat $\euclid^{12}$ space, $\chordalkernel$ acts as a Euclidean ARD kernel and inherits its characteristic property. However, the chordal distance underestimates true geodesic displacement (see \figref{fig:configuration-distances}). The second kernel $\geodesickernel$ is a geodesic exponential kernel
\begin{equation}
\label{eq:geodesic-kernel}
    \geodesickernel(\jointfiber, \altjointfiber) = \prod_{k \in \{1, 2\}} \exp\left( -\frac{1}{2}\left\| \log(\inv{\fiber_{t_k}}\altfiber_{t_k}) \right\|_{\inv{\Gamma}_k}^2\right),
\end{equation}
with $\gamma_k \in \euclid_+^6$. This kernel is \textit{not} characteristic due to the manifold curvature and the presence of a cut locus \cite{feragen2015geodesic}. However, \eqref{eq:geodesic-kernel} recovers local characteristic behavior provided the marginal distributions are concentrated and close to one another, and the kernel is sharp. To show this, suppose that the distributions $\probmeasure_{\fiber_t}$ and $\probmeasure_{\altfiber_t}$ are concentrated, and that $\probmeasure_{\fiber_t}$ is close to $\probmeasure_{\altfiber_t}$ at both $t_1$ and $t_2$ (so that $\log$ remains well-defined and injective).

\begin{definition}
    A probability distribution $\probmeasure_{\fiber_t}$ is \textit{concentrated} if its support is in the domain of the logarithm map centered at the distribution mean $\meanfiber_t$, i.e., $\text{supp}(\probmeasure_{\fiber_t}) \subseteq \text{dom}(\log_{\meanfiber_t})$.
\end{definition}

\begin{definition}
    Two distributions $\probmeasure_{\fiber_t}$, $\probmeasure_{\altfiber_t}$ are \textit{close} if the intersection of their supports is in the domain of the logarithm map centered at the observed distribution mean $\altmeanfiber_t$, i.e., $\text{supp}(\probmeasure_{\fiber_t}) \cup \text{supp}(\probmeasure_{\altfiber_t}) \subseteq \text{dom}(\log_{\altmeanfiber_t})$.
\end{definition}

\begin{figure}
    \centering
    \includegraphics[width=\linewidth]{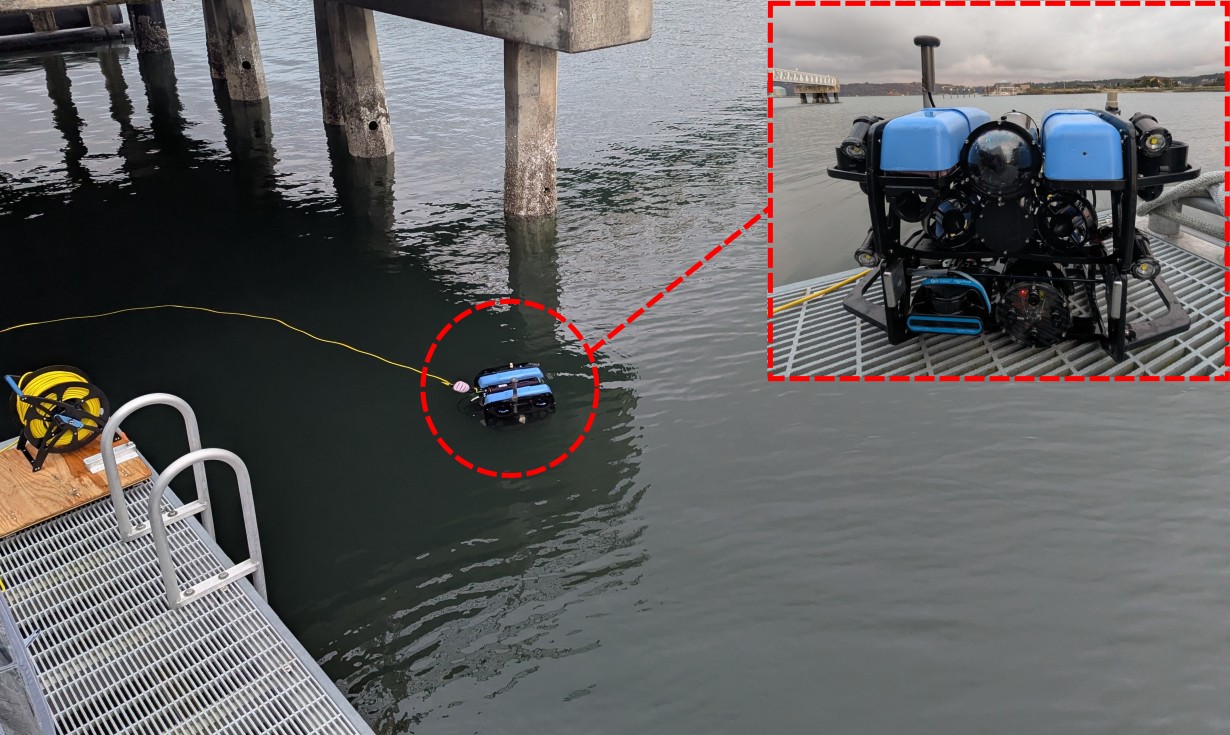}
    \caption{We performed our hardware experiments at a ship dock using a modified, tethered BlueROV2 Heavy \cite{bluerov}. The vehicle navigated through dock pylons using model predictive path integral (MPPI) control and the learned geometric SPINNs.}
    \label{fig:bluerov2}
\end{figure}

Under the concentration and closeness conditions, the geodesic distance can be analyzed using the Baker-Campbell-Hausdorff (BCH) series expansion. Letting $\fiberexp = \log(\fiber)$ and $\altfiberexp = \log(\altfiber)$, the expansion yields
\begin{equation}
\label{eq:bch}
\begin{split}
    \log(\inv{\fiber}\altfiber) \oslash \gamma \approx 
    &\left( \altfiberexp - \fiberexp \right. \\
    &-\frac{1}{2}\left[ \fiberexp, \altfiberexp \right] \\
    &+\frac{1}{12}\left( \left[ \fiberexp, \left[ \fiberexp, \altfiberexp \right] \right] -\left[ \altfiberexp, \left[ \altfiberexp, \fiberexp \right] \right] \right) \\
    &+\left.\dots \right) \oslash \gamma,
\end{split}
\end{equation}
where $\oslash$ denotes the Hadamard quotient and $[\cdot, \cdot]$ is the Lie bracket. The higher-order terms in \eqref{eq:bch} capture the non-Euclidean curvature of $\SEthree$. If the bandwidths $\gamma$ are chosen to be sufficiently small, the kernel becomes sharp, causing these higher-order terms to vanish faster than the linear displacement.

\begin{definition}
    The squared geodesic exponential kernel is \textit{sharp} if the bandwidths $\gamma$ are sufficiently small such that the higher-order terms of the BCH series \eqref{eq:bch} are negligible and the geodesic distance is approximately Euclidean.
\end{definition}

\begin{proposition}
    For concentrated and close joint marginals $\jointfiberlaw$ and $\altjointfiberlaw$, a sharp geodesic exponential kernel satisfies
    \begin{equation}
        \geodesickernel(\jointfiber, \altjointfiber) \approx \prod_{k \in \{1, 2\}} \exp\left( -\frac{1}{2}\left\| \log(\altfibercirc_{t_k}) - \log(\fibercirc_{t_k}) \right\|_{\inv{\Gamma}}^2\right).
    \end{equation}
    In this regime, the geodesic exponential kernel behaves as an ARD kernel on $\sethree$, ensuring the kernel score acts locally as a strictly proper scoring rule.\footnotemark\
\end{proposition}

\subsubsection{Training Objective}

The velocity and configuration losses are defined using their corresponding scoring rules over the predicted and observed trajectories,
\begin{align}
    \loss_{\text{vel}}(\neuralparams) &= \sum_{n=1}^{N-\ell} \scorerule \Bigl( \probmeasure_{(\fibercirc_{t_n}, \fibercirc_{t_{n+\ell}})}, \bigl( \altfibercirc_{t_n}, \altfibercirc_{t_{n+\ell}} \bigr) \Bigr),\label{eq:velocity-fdm} \\[3pt]
    \loss_{\text{config}}(\neuralparams) &= \sum_{n=1}^{N-\ell} \scorerule \Bigl( \probmeasure_{(\fiber_{t_n}, \fiber_{t_{n+\ell}})}, \left( \altfiber_{t_n}, \altfiber_{t_{n+\ell}} \right) \Bigr),\label{eq:configuration-fdm}
\end{align}
where $\ell$ controls the time separation between discrete evaluation points. The FDM training objective is the expected sum of these losses:
\begin{equation}
    \loss_{\text{fdm}}(\neuralparams) = \mathbb{E}_{s} \left[ \loss_{\text{vel}}(\neuralparams) + \loss_{\text{config}}(\neuralparams) \right].
\end{equation}

While $\loss_{\text{fdm}}$ is theoretically sufficient for learning the stochastic dynamics of underwater vehicles, we found that it struggles to converge in field applications where the dynamics prior is inaccurate and uncertainty is high. Under these circumstances, FDM can produce stiff models that fail to capture the underlying process. We address this issue by augmenting the objective with a mean squared error term:
\begin{equation}
    \loss_{\text{mse}}(\neuralparams) = \sum_{p=1}^P\sum_{n=1}^N \| \log(\inv{\fiber_{p, n}}\altfiber_n^p) \|^2 + \| \altfibercirc_n^p - \fibercirc_n^p \|^2.
\end{equation}
The resulting composite training objective is described by
\begin{equation}
    \loss_{\text{ocean}}(\neuralparams) = \mathbb{E}_{s} \left[ \loss_{\text{vel}}(\neuralparams) + \loss_{\text{config}}(\neuralparams) + \loss_{\text{mse}}(\neuralparams) \right].
\end{equation}

\footnotetext{In practice, the requirements for kernel sharpness and distribution concentration are relatively mild. Concentrated distributions on Lie groups are widely-used \cite{chirikjian2012stochastic, ye2025uncertainty, wang2008nonparametric, long2013banana, barfoot2014associating, barfoot2024state}, and we found that the range of bandwidths over which the Euclidean approximation holds are sufficiently large for training. A more significant challenge lies in ensuring that the predicted and observed joint marginals are close. This problem can be addressed by pre-training the drift $\neuraldrift$ or embedding strong dynamical priors into the model architecture. We also found that augmenting the training objective with a mean squared error (MSE) loss term into the training is a practical solution when a strong prior is unavailable. The MSE loss steers the model toward the high-probability regions where the geodesic kernel score recovers its strictly proper behavior.}
\section{Experimental Results}\label{sec:experiments}

We evaluate the effectiveness of our geometric SPINN for learning stochastic underwater vehicle dynamics using a simulated underwater vehicle and a modified BlueROV2 Heavy \cite{bluerov} operating in a harbor environment. The proposed and baseline models are implemented using \texttt{JAX} \cite{jax2018github} and \texttt{Equinox} \cite{kidger2021on}. The MLPs used in the experiments are configured with 3 hidden layers and 32 neurons per layer. We train all models using the Adam optimizer \cite{kingma2014adam} with a learning rate of $0.0001$. The code used for our experiments will be made publicly available online upon paper acceptance.

\begin{figure*}
    \centering
    \includegraphics[width=\linewidth]{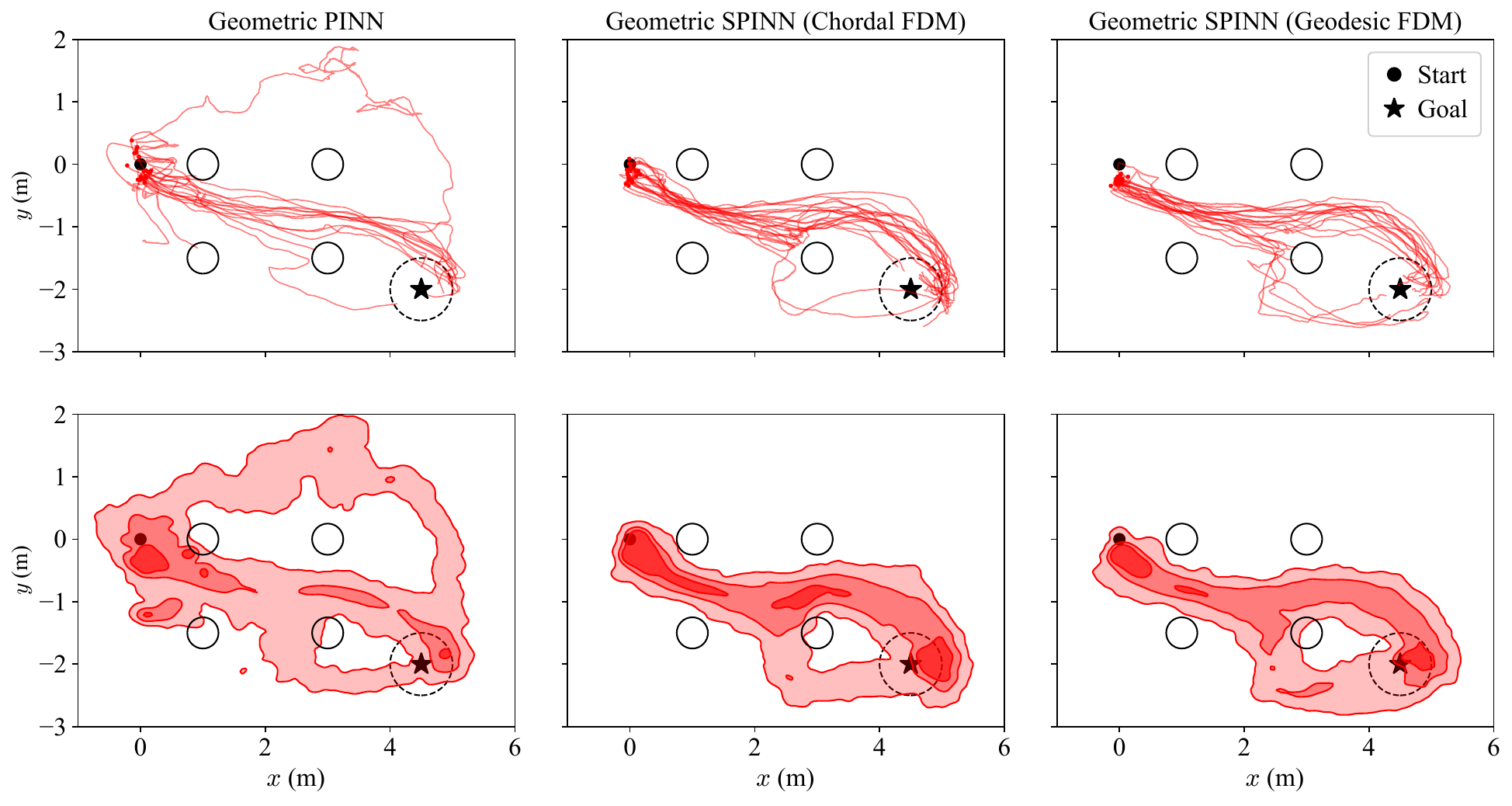}
    \caption{Field experiment results. The start position is set to $(0, 0, -1.5)$~m, and the goal position is $(4.5, -2.0, -1.5)$~m. Variance in the starting configuration is caused by controller steady-state error and tide changes; experiments were conducted during slack tide and flood tide. The top panel depicts vehicle trajectories executed under MPPI control using the baseline geometric PINN (13/20 success) and the proposed stochastic geometric SPINNs trained with chordal FDM (17/20 success) and geodesic FDM (18/20 success). The bottom panel displays contour maps representing the search space considered by each MPPI controller during execution, where darker shades represents regions with high sample density. The high-density sample regions explored under the Geometric PINN are tight while the sample regions of the same density under the stochastic models are more distributed.}
    \label{fig:ocean-results}
\end{figure*}

\subsection{Simulation Experiments}

We first verified our stochastic dynamics learning approach through an ablation study in simulation, where we simulated the dynamics of an underwater vehicle subject to environmental disturbances. The study was split into two steps. First, we evaluated the models in a low-complexity, data-rich regime to establish performance baselines. Second, we evaluated the models in a moderate-complexity, sparse data regime representative of the expected hardware deployment conditions.

Two training datasets were collected, accordingly. The first dataset was constructed from a single control scenario comprising 100 ten-second trajectories generated under a constant forward thrust command. The second dataset spanned 10 control scenarios, each containing 20 ten-second trajectories, where the vehicle was driven by a distinct feedforward command (see \figref{fig:maneuvers}). The control maneuvers were chosen to cover typical underwater vehicle control actions. All trajectories were sampled at 20~Hz and post-processed using a 4~s window. The trajectories were normalized so that each began at the identity configuration $\groupid$, ensuring the models learned the local dynamics of the system.

The models implemented in our ablation study include: a geometric PINN with identical architecture to our geometric SPINN but without the diffusion network (similar to \cite{duong2024port}) and a Euclidean SPINN with identical architecture to our geometric SPINN but with Euclidean kinematic constraints. The Euclidean SPINNs were trained using the distance-aware diffusion objective in \cite{djeumou2023how}, the unconstrained distance-aware diffusion objective in \cite{koprulu2025neural}, finite dimensional matching via our velocity objective \eqref{eq:velocity-fdm} and an ARD kernel score for the Euclidean configuration \cite{zhang2025efficient}, and moment matching. Because the prior work has demonstrated the superior performance of physics-informed methods over unstructured, non-parametric methods \cite{duong2024port, djeumou2023how}, we focus on comparison with other physics-informed methods. We pretrained the hydrodynamic coefficient networks \eqref{eq:hydrodynamic-networks} so that the predicted matrices were within the expected coordinate chart of the true dynamics.\footnotemark\ The models were assessed by generating 350 ten-second samples for each control scenario from the true and predicted models. We report the mean-squared error and Wasserstein 2-distance averaged across the full time horizon, and repeat the experiment over 5 seeds.

\footnotetext{The generalized coordinates of underwater vehicles are heterogeneous. As a result, training models from scratch is challenging because it requires learning the geometry of the coordinate chart, which can involve predicting values that are separated by multiple orders of magnitude. Pre-training the models to match the expected coordinates mitigates this issue. In future work, coordinate-independent identification methods \cite{yang2026coordinate} can be used instead.}

\begin{figure*}
    \centering
    \includegraphics[width=\linewidth]{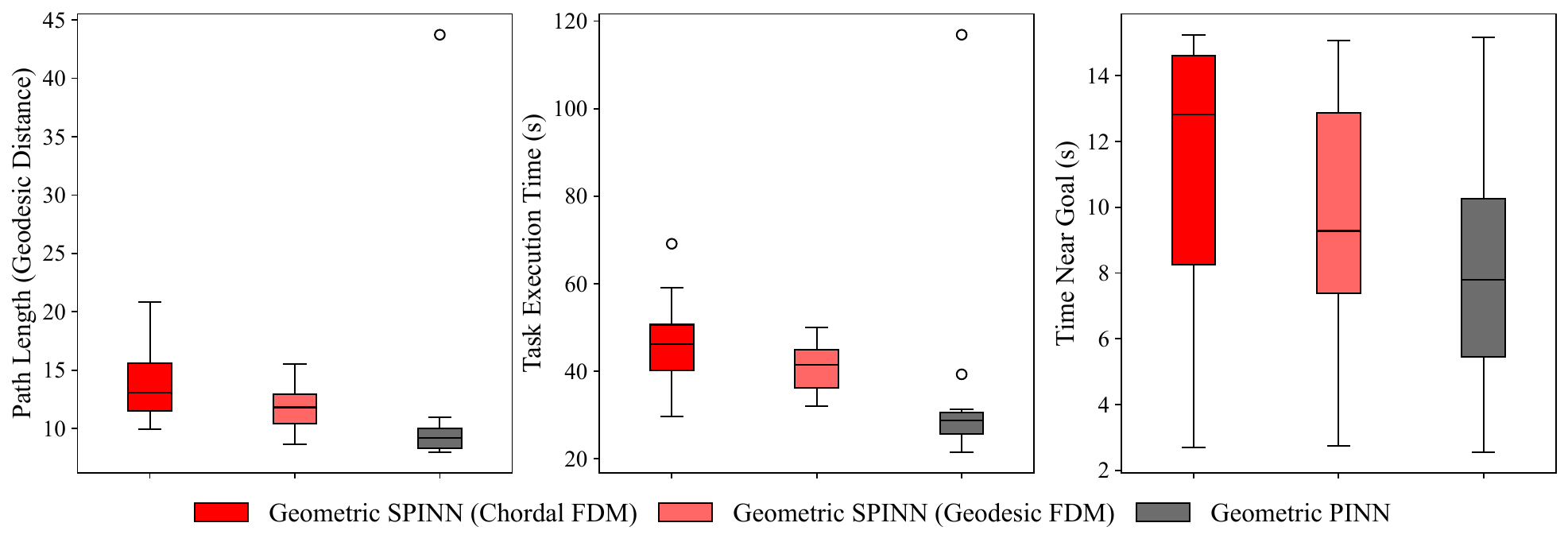}
    \caption{Path statistics for successful task instances. The ``Time Near Goal'' metric represents the amount of time that the vehicle spent within 1~m of the goal during task execution. The results demonstrate that the proposed methods prefer longer, risk-aware paths, whereas the deterministic baseline relies on aggressive, collision-prone maneuvers (see \figref{fig:ocean-results}).}
    \label{fig:ocean-stats}
\end{figure*}

The simulation results presented in \figref{fig:simulation-results} and \figref{fig:simulation-distributions} demonstrate that the proposed geometric SPINNs trained under the chordal and geodesic FDM strategies match the true trajectory distributions more accurately than all baselines, while maintaining competitive or superior performance in terms of RMSE. This performance gain was most notable in the sparse data regime, where the geometric inductive biases improve sample efficiency and physical consistency. We found that the velocity FDM training objective \eqref{eq:velocity-fdm} enabled the largest gain in distribution matching performance, while other SPINN training strategies struggled to capture the system's inherent aleatoric uncertainty, instead collapsing onto near-deterministic solutions regardless of geometric architectural constraints. This finding suggests that, while geometric inductive biases can improve distribution matching, they are independently insufficient for enabling accurate stochastic model learning. 

The simulation experiments revealed an optimization pathology when training the Euclidean SPINN via FDM: the discontinuous state representation destabilizes training, leading to frequent training failures ($60\%$ training success rate across 5 seeds). Other Euclidean baselines bypass this issue because their low-variance, near-deterministic behavior avoids these unstable regions of the state space. In contrast, our geometric SPINN formulation yields stable training dynamics, achieving a $100\%$ training success rate across all training strategies. Finally, we note that the Euclidean SPINN trained via the unconstrained distance-aware diffusion objective \cite{koprulu2025neural} failed to train in both data regimes--which is caused by instability in the interaction between the diffusivity matrix and mass matrix under the negative log-likelihood objective--so we declined to document this method in our results.

\subsection{Field Experiments}

We tested our proposed geometric SPINN trained with chordal FDM and geodesic FDM on hardware in an obstacle-avoidance navigation task representative of marine infrastructure inspection. We compared our methods against the geometric PINN to evaluate the performance gap between stochastic and deterministic approaches. We did not deploy the Euclidean SPINN because its training was either unstable or converged to an almost-deterministic solution. The navigation problem was formulated as a stochastic trajectory optimization problem,
\begin{subequations}
\label{eq:optimal-control}
\begin{alignat}{2}
    \min_{u}& \hspace{2mm} && \mathbb{E} \left[ \int_0^T \left\|\log\left(\inv{\fiber}\fiber^{\text{goal}}\right) \right\|_L^2 + \left\|\fibercirc^{\text{goal}} - \fibercirc \right\|_Q^2 \right] \\[2pt]
    \text{s.t.}& && \dd (\fiber, \fibercirc) = \neuraldrift(\fiber, \fibercirc, \control)\,\dt + \neuraldiffusion(\fibercirc) \circ \dw \\
    & && \probmeasure(\fiber \in \mathcal{G}) \geq 1 - \alpha \\
    & && u \in \mathcal{U}
\end{alignat}
\end{subequations}
where $(\fiber^{\text{goal}}, \fibercirc^{\text{goal}})$ is the goal state, $\mathcal{G}$ represents the the collision-free configuration space, and $\alpha \in (0, 1)$ denotes the collision-avoidance risk threshold. We implemented \eqref{eq:optimal-control} using model predictive path integral (MPPI) control \cite{williams2016aggressive}.

The hardware experiments were conducted with a tethered BlueROV2 Heavy at \censor{the Oregon State University Hatfield Marine Science Center}, located on \censor{the Yaquina Bay in Newport, Oregon} (see \figref{fig:bluerov2}). All data collection and tests were performed during slack tide and flood tide. The primary disturbances included wind waves, ship wake, and time-varying currents ranging $0-0.5$~m/s. The testing conditions were selected based on the operating limits of the vehicle. 

The BlueROV2 was equipped with an IMU (sampled at $100\,\mathrm{Hz}$) and a Nortek Nucleus 1000 DVL (sampled at $5\,\mathrm{Hz}$) \cite{dvl}. State estimates were obtained from an extended Kalman filter (EKF) at a rate of $20$~Hz. We used dead reckoning for our position estimates because we did not have access to a positioning sensor. All state estimation and low-level control software was run on a Raspberry Pi 5 located onboard the vehicle, while the trained models and MPPI controller were deployed from a top-side laptop computer with a GeForce RTX 2070.

We collected a training dataset comprising 10 control scenarios, each with 20 ten-second trajectories. For each trajectory, a proportional-derivative controller drove the vehicle to an initial configuration; upon reaching steady state, the vehicle transitioned into open-loop feedforward control with constant input (see \figref{fig:maneuvers}). The final dataset totaled about $30$ minutes of data, post-processed with a $4$~s window and normalized to begin at the identity configuration $\groupid$.

The MPPI controller ran at a rate of $5$~Hz, generating $1000$ samples at each step. The optimal control sequence and trajectory were sub-sampled at $20$~Hz using (geodesic) linear interpolation, then tracked by an impedance controller
\begin{equation}
    \control = \control^{\text{opt}} + K_p \left(\log(\inv{\fiber}\fiber^{\text{opt}})\right) + K_d \left( \fibercirc^{\text{opt}} - \fibercirc \right)
\end{equation}
at a rate of $50$~Hz, where $\control^{\text{opt}}$ and $(\fiber^{\text{opt}}, \fibercirc^{\text{opt}})$ denote the optimal control and state, respectively.

The learned baseline and proposed models were evaluated in the navigation task across 20 consecutive trials. A trial was considered successful if the vehicle moved within a 0.5~m radius of the goal without obstacle collisions. Trials were terminated early for unsafe behavior (e.g., uncontrolled spinning), failure to make progress, or if the vehicle remained within a 1~m goal radius for more than 15~s without entering the target zone. Additional demonstrations in calm and choppy water conditions are presented in our supplementary video.

\figref{fig:ocean-results} displays the executed trajectories. The proposed geometric SPINN trained via geodesic FDM achieved the highest success rate (18/20 trials), followed by the chordal FDM variant (17/20 trials) and the geometric PINN (13/20 trials). The geometric PINN struggled with hardware transfer, and control behavior under the model was aggressive compared to the geometric SPINNs (see \figref{fig:ocean-stats}), causing collisions and unsafe behavior. The chordal FDM model exhibited cautious behavior avoiding collisions at the expense of longer execution times and three task failures from lack of progress near the goal. Conversely, the geodesic FDM variant exhibited less cautious control behavior, but collided twice with the dock pylons.

\section{Conclusion}\label{sec:conclusion}

This paper presents a novel method for learning underwater vehicle dynamics using stochastic physics-informed neural networks (SPINNs). Our approach integrates Euler-Poincar\'{e} dynamics and Lie group constraints as inductive biases within the architecture of a SPINN, and utilizes geometrically-consistent training strategies based on moment matching and finite dimensional matching. In simulation, the proposed framework yields a more stable training process and improved prediction performance compared to the baseline methods. Furthermore, field experiments demonstrate the merit of modeling stochastic dynamics in underwater navigation, showing that model predictive control algorithms that explicitly account for stochasticity produce safer actions than those that neglect system uncertainty.

Future work includes exploring realistic disturbance models (e.g., anisotropic, multi-modal, or mean-reverting noise) that better represent disturbances in natural environments. Another intriguing direction for future work is to extend our stochastic dynamics-learning framework to stochastic variational integrators \cite{bourabee2009stochastic} to enable stable, long-horizon underwater motion planning. Finally, scaling our formulation to multi-body dynamics may better facilitate deployment of autonomous floating-base manipulator systems to oceanic environments.
\appendices

\section{}
\label{app:integrator}

We integrate the proposed geometric SPINN (Section~\ref{subsec:architecture}) using the partitioned integrator described in Algorithm~\ref{alg:integrator}. A Lie-Euler step \cite{iserles2000lie} is used for the configuration evolution to prevent drift from group constraints, and an Euler-Heun step \cite{roberts2012modify} is used for the velocity evolution to preserve the Stratonovich interpretation of the stochastic integral. The diffusion midpoint approximation can be dropped altogether to integrate deterministic systems. We implement the integrator using \texttt{JAX} \cite{jax2018github} and \texttt{Diffrax} \cite{kidger2021on}.
\begin{algorithm}
    \caption{Lie Group Integrator}\label{alg:integrator}
    \SetKwInOut{Input}{Input}
    \Input{Model parameters $\neuralparams$, current state $(\fiber_k, \fibercirc_k)$, control input $\control_k$, time increment $\timeinc$, Brownian increment $\brownianinc$}\vspace{5pt}
        $\fibercirc_k' \gets \inv{M}_\neuralparams(\fibercirc_k) \left( \dadj_{\fibercirc_k}\genmomcirc_k + \lieforce_\neuralparams(\fiber_k, \fibercirc_k, \control_k) \right)\,\timeinc$\;\vspace{3pt}
        $\metricdiffusion_k \gets \metricdiffusion_{\neuralparams}(\fibercirc_k)\,\brownianinc$\;\vspace{3pt}
        $\metricdiffusion'_k \gets \metricdiffusion_{\neuralparams}(\fibercirc_k + \metricdiffusion_k)\,\brownianinc$\;\vspace{3pt}
        $\fiber_{k+1} \gets \fiber_k \exp ( \fibercirc_k\,\timeinc )$\;\vspace{3pt}
        $\fibercirc_{k+1} = \fibercirc_k + \fibercirc'_k + \frac{1}{2}(\metricdiffusion_k + \metricdiffusion'_k)$\;
\end{algorithm}

\bibliographystyle{IEEEtran}
\bibliography{references}

\end{document}